\documentclass[11pt]{article}
\usepackage{paclic33}
\usepackage{latexsym}
\usepackage{amsmath}
\usepackage{multirow}
\usepackage{url}
\usepackage{float}
\usepackage{pgfplots}
\usepackage{footnote}
\usepackage[flushleft]{threeparttable}
\usepackage[utf8]{inputenc}
\usepackage[T5]{fontenc}
\usepackage{longtable}
\usepackage{supertabular}
\usepackage{makecell}
\usepackage{multicol}
\usepackage{graphicx}
\usepackage{array}
\usepackage{lipsum}

\DeclareUnicodeCharacter{0301}{\'{e}}
\newcommand{\footref}[1]{%
    $^{\ref{#1}}$%
}
\newcolumntype{L}[1]{>{\raggedright\let\newline\\\arraybackslash\hspace{0pt}}m{#1}}
\newcolumntype{C}[1]{>{\centering\let\newline\\\arraybackslash\hspace{0pt}}m{#1}}
\newcolumntype{R}[1]{>{\raggedleft\let\newline\\\arraybackslash\hspace{0pt}}m{#1}}

\makeatletter
\@ifpackageloaded{hyperref}{%
  \renewcommand\footref[1]{%
    \begingroup 
    \unrestored@protected@xdef\@thefnmark{%
      \ref*{#1}%
    }%
    \endgroup 
    \ifHy@hyperfootnotes 
       \expandafter\@firstoftwo 
    \else 
       \expandafter\@secondoftwo 
    \fi 
    {\hyperref[#1]{\strut\H@@footnotemark}}{\@footnotemark}%
  }%
}{}%

\newcommand\savedlabel{}%
\AtBeginDocument{\let\savedlabel=\label}%
\newcommand\footnotereflabel[1]{%
   \@bsphack
   \begingroup
   \def\@currentHref{Hfootnote.\theHfootnote}\savedlabel{#1}%
   \endgroup
   \@esphack
}%

\makeatother

\title{Empirical Study of Text Augmentation on Social Media Text in Vietnamese}

\author{Son T. Luu \\
  University of Information Technology \\
  VNU-HCM, Vietnam \\
  {\tt sonlt@uit.edu.vn} \\  \And
  Kiet Van Nguyen \\
  University of Information Technology \\
  VNU-HCM, Vietnam \\
  {\tt kietnv@uit.edu.vn} \\ \AND
  \textbf{Ngan Luu-Thuy Nguyen} \\
  University of Information Technology \\
  VNU-HCM, Vietnam \\
  {\tt ngannlt@uit.edu.vn} \\
  \newline
}

\date{}

\begin{document}
\maketitle
\begin{abstract}
In the text classification problem, the imbalance of labels in datasets affect the performance of the text-classification models. Practically, the data about user comments on social networking sites not altogether appeared - the administrators often only allow positive comments and hide negative comments. Thus, when collecting the data about user comments on the social network, the data is usually skewed about one label, which leads the dataset to become imbalanced and deteriorate the model's ability. The data augmentation techniques are applied to solve the imbalance problem between classes of the dataset, increasing the prediction model's accuracy. In this paper, we performed augmentation techniques on the VLSP2019 Hate Speech Detection on Vietnamese social texts and the UIT - VSFC: Vietnamese Students' Feedback Corpus for Sentiment Analysis. The result of augmentation increases by about 1.5\% in the F1-macro score on both corpora.
\end{abstract}

\section{Introduction}
\label{introduction}
In recent years, the growth of hate speech has become a crime, not only face-to-face action but also online communication  \cite{10.1145/3232676}. The development of social network nowadays had made this situation worse. The threading of harassment comments and harassment speech makes the user stop expressing their opinions and looking up for other ideas \cite{sonvx2019}. Fortuna and Nunes \shortcite{10.1145/3232676} defined hate speech as the language attacking, diminishing, and inciting violence or hate against individuals or groups based on their characteristics, religion, nations, and genders. To solve this problem, many datasets are constructed to detect and classify user comments on social network sites such as Facebook, Twitter, and Facebook in many languages\footnote{\url{http://hatespeechdata.com/}}. 

The HSD-VLSP dataset \cite{sonvx2019} provided by the VLSP 2019 Shared task about Hate speech detection on Social network\footnote{\url{https://www.aivivn.com/contests/8}} contained nearly 25,000 comments and posts of Vietnamese Facebook users and has three labels. However, the distribution of three classes in the dataset is imbalanced. Besides, the UIT-VSFC dataset \cite{8573337} that was used for predicting the feedback from students contained about 16,000 sentences and was annotated for two different tasks: sentiment analysis and topic analysis. Same as the HSD-VLSP dataset, the distribution of labels on the UIT-VSFC dataset is also imbalanced. We use the data augmentation techniques to generate new comments that belong to minority classes from the original dataset to tackle those restrictions. We conduct experiments on the augmented dataset and compare it with the original dataset to indicate data augmentation effectiveness. Those augmentation techniques include synonym replacement, random insertion, random swapping, and random deletion \cite{wei-zou-2019-eda}.

The rests of the paper are structures as below. Section \ref{related_works} introduces recent works in hate speech detection. Section \ref{dataset} gives an overview of two datasets include the HSD-VLSP dataset and the UIT-VSFC dataset. Section \ref{methodologies} presents the methods and models used in our paper. Section \ref{empirical_results} shows our experiment results when applied to the text augmentation techniques. Section \ref{conclusion} concludes the paper. 
\section{Related works}
\label{related_works}
Duyen et al. \shortcite{7043403} conducted an empirical study about the sentiment analysis for Vietnamese texts based on machine learning to study the influences on the models' accuracy. However, besides the impact of the model's ability, and the feature selection such as word-based, syllable-based, and extracting essential words, the imbalance in the dataset also affects the result. The imbalance in label distribution happens regularly \cite{Ali2015ClassificationWC} when one class seems to be more interested than the other. For example, in social media networks, the abusive and hateful comments are often hidden by the users or administrators, since the clean comments take the majority part. The VLSP2019 hate speech dataset \cite{sonvx2019} and the UIT-VSFC dataset \cite{8573337} also suffer the imbalance in class distribution. The detail of those datasets is showed in Section \ref{methodologies}.

Wang and Yang \shortcite{wang-yang-2015-thats} provided a novel method for enhancing the data used for behavior analysis using social media texts on Twitter. Their approaches include using the lexical embedding and frame-semantic embedding. The obtained results showed that using the data augmentation brings significantly better results than no data augmentation (using Google New Lexical embedding brings 6.1\% improvement in F1-score and using additional frame-semantic embedding from Twitter brings 3.8\% improvement in F1-score.

Ibrahim et al. \shortcite{ibrahim2018} presented different data augmentation techniques for solving the imbalance problem in the Wikipedia dataset and an ensemble method used for the training model. The result achieved a 0.828 F1-score for toxic and nontoxic classification, and 0.872 for toxicity types prediction. 

Rizos et al. \shortcite{10.1145/3357384.3358040} introduced data augmentation techniques for hate speech classification. The authors 's proposed methods increased the result of hate speech classification to 5.7\% in F1-macro score. 

Finally, Wei and Zou \shortcite{wei-zou-2019-eda} provided EDA (Easy Data Augmentation) techniques used to enhance data and boost performance on the text classification task. It contains four operations: synonym replacement, random insertion, random swap,  and random deletion. In this paper, these operations are applied to the HSD-VLSP 2019 dataset and the UIT-VSFC dataset to increase the classification models' ability.

\section{Datasets}
\label{dataset}
\subsection{The HSD-VLSP dataset}
\label{vlsp:dataset}
The hate speech dataset was provided by the VLSP 2019 shared task about hate speech detection for social good \cite{sonvx2019}. The dataset contains a total of 20,345 comments and posts crawled from Facebook. Each comment is labeled by one of three labels: CLEAN, OFFENSIVE, and HATE. Table \ref{tab:dataset_overview} showed the overview information about the dataset. 

\begin{table}[H]
    \begin{center}
        \begin{tabular}{|p{2.2cm}|R{1.3cm}|R{1.3cm}|R{1.4cm}|}
            \hline
        \textbf{} & \textbf{Num. comments} & \textbf{Avg. word length} & \textbf{Vocab. size} \\
        \hline
        CLEAN & 18,614 & 18.69 & 347,949 \\
        \hline
        OFFENSIVE & 1,022 & 9.35 & 9,556   \\
        \hline
        HATE & 709 & 20.46 & 14,513 \\
        \hline
        Total & 20,345 & 18.28 & 372,018 \\
        \hline
        \end{tabular}
    \end{center}
    \caption{Overview of the HSD-VLSP dataset}
    \label{tab:dataset_overview}
\end{table}

According to Table \ref{tab:dataset_overview}, the number of CLEAN comments take a majority part in the dataset, the number of OFFENSIVE comments and HATE comments are much fewer. Thus, the distribution of labels in the dataset is imbalanced.

\subsection{The UIT-VSFC dataset}
The Vietnamese Students’ Feedback Corpus for Sentiment Analysis (UIT-VSFC) by Nguyen et al. \shortcite{8573337} are used to improve the quality of education. The dataset contains nearly 11,000 sentences and consists of two tasks: sentiment-based classification and topic-based classification. The sentiment-based task comprises three labels: positive, negative, and neutral. The topic-based task comprises four labels corresponding to lecturer, training program, facility, and others. Table \ref{tab:vsfc_overview_train} describes the overview about the UIT-VSFC training set.

\begin{table}[H]
    \begin{center}
        \begin{tabular}{|p{2.1cm}|R{1.3cm}|R{1.3cm}|R{1.4cm}|}
            \hline
                \textbf{} & \textbf{Num. comments} & \textbf{Avg. word length} & \textbf{Vocab. size} \\
                \hline
                Total & 11,426 & 10.2 & 117,295 \\
                \hline
                \multicolumn{4}{|c|}{\textbf{Sentiment based task}}\\
                \hline
                Positive & 5,643 & 8.2 & 46,807 \\
                \hline
                Negative & 5,325 & 12.6 & 67,193  \\
                \hline
                Neutral & 458 & 7.1 & 3,295 \\
                \hline
                \multicolumn{4}{|c|}{\textbf{Topic based task}} \\
                \hline
                Lecturer & 8,166 & 9.7 & 79,854 \\
                \hline
                Training program & 2,201 & 12.2 & 27,039 \\
                \hline
                Facility & 497 & 12.3 & 6,130 \\
                \hline
                Others & 562 & 10.9 & 4,272 \\
                \hline
        \end{tabular}
    \end{center}
    \caption{Overview of the UIT-VSFC training set}
    \label{tab:vsfc_overview_train}
\end{table}

According to Table \ref{tab:vsfc_overview_train}, the number of data in the neutral label is lower than positive and negative on the sentiment-based task. So is the topic-based task when the \textit{facility} and \textit{others} labels are much lower than the two remain labels. In brief, the imbalance data happened on the neutral label for the sentiment-based task, and the \textit{facility} and the \textit{other} labels for the topic-based task.

\section{Our proposed method}
\label{methodologies}

\subsection{The augmentation techniques}
In this paper, we implement the EDA techniques introduced by Wei and Zou \shortcite{wei-zou-2019-eda}. Those techniques will get a sentence as input and perform one of these following operations to generate new comments:

\begin{itemize}
\item \textbf{Synonym replacement (SR)}: This operation creates a new sentence by randomly choosing $n$ words from the input sentence and replaces them by their synonyms, excluding the stop words. In our experiments, we use the Vietnamese wordnet\footnote{\footnotereflabel{vi-wordnet} \url{https://github.com/zeloru/vietnamese-wordnet}} from Nguyen et al. \shortcite{nguyen2016two} for synonym replacement and the Vietnamese stopword dictionary\footnote{\url{https://github.com/stopwords/vietnamese-stopwords}} for removing stop words in the sentence.

\item \textbf{Random Insertion (RI)}: This operation generates new data by first finding a random word in the input sentence, which is not a stop word, then taking its synonym and putting it into the sentence's random position. The synonyms are taken from the Vietnamese wordnet\footref{vi-wordnet}.

\item \textbf{Random  Swap  (RS)}: This operation makes a new sentence by choosing two random words in the input sentence and swap their position.

\item \textbf{Random  Deletion  (RD)}: This operation creates a new sentence by accidentally deleting $p$ words in the sentence ($p$ is the probability defined before by the user).
\end{itemize}

According to Wei and Zou \shortcite{wei-zou-2019-eda}, $n$ indicates the number of changed words for SR, RI, and RS methods, which calculated as $n=\alpha*l$, where $\alpha$ is the percentage of replacement word in the sentence and $l$ is the length of the sentence. For the RD method, the probability of deletion words $p$ equal to $\alpha$. The $\alpha$ is defined by the user.

Table \ref{tab:eda_example} shows examples of data between original and after augmented by EDA techniques in the HSD-VLSP dataset. 

\begin{table}[H]
    
    \begin{center}
        \begin{tabular}{|p{5.9cm}|c|}
        \hline
        \textbf{Comments} & \textbf{Type} \\
        \hline
        
        \makecell[l]{\textbf{Original}: con này xấu trai vl \\ \textit{(this guy is f*cking ugly)} \\ \textbf{Augmented}: con xấu trai vl} & RD\\
        \hline
        
        \makecell[l]{\textbf{Original}: Đcm nản vl \\ \textit{(This is f*cking bored)} \\ \textbf{Augmented}: Đcm nhụt chí vl} & SR\\
        \hline
        
        \makecell[l]{\textbf{Original}: Đume đau răng vl \\ \textit{(Toothache got damn hurt!)} \\ \textbf{Augmented}: Đume răng đau vl} & RS\\
        \hline
        
        \makecell[l]{\textbf{Original}: Đm Lắm chuyện vl \\ \textit{(F*uck those curious guys)} \\ \textbf{Augmented}: \\Đm thứ Lắm chuyện vl} & RI\\
        \hline
        \end{tabular}
    \end{center}
    \caption{Several example of the augmented data on the HSD-VLSP dataset}
    \label{tab:eda_example}
\end{table}

\subsection{The classification model}
Aggarwal and Zhai \shortcite{books/sp/mining2012/AggarwalZ12b} defined the text classification problem as a set of training data $D=\{X_1, ..., X_N\}$, in which each record is labeled with a class value drawn from a set of discrete classes indexed by $\{1..k\}$. The training data used to construct a classification model. With a given test dataset, the classification model is used to predict a class for each instance in the test dataset. Our paper used the Text-CNN model \cite{kim-2014-convolutional} for the HSD-VLSP dataset and the Maximum Entropy model \cite{nigam1999using} for the UIT-VSFC dataset to study the effectiveness of data augmentation on those two datasets. In practice, the idea of Logistic Regression is maximizing the cross-entropy loss of the actual label in the training dataset \cite{jurasky2000speech}, which is the same as the Maximum Entropy model \cite{nigam1999using}. Thus, we use the term Maximum Entropy instead of Logistic Regression in our results.
\section{Empirical results}
\label{empirical_results}

\subsection{Experiment configuration}
For the HSD-VLSP corpus, we use cross-validation with five folds for the Text-CNN model and the Maximum Entropy model. Following the same manner in the previous study \cite{luu2020comparison}, for each fold, we keep the test set and enhance the training set with EDA techniques.

For the UIT-VSFC dataset, we used the data divided into the training, development, and test sets by Nguyen et al. \shortcite{8573337}. Then we run the EDA techniques on the training set and use the test set to evaluate the result. 

\subsection{Data augmentation result}
We first applied the EDA techniques on the entire original HSD-VLSP dataset to enhance the data on HATE and OFFENSIVE labels. Table \ref{tab:dataset_augmented} describes the information about the HSD-VLSP dataset after making data augmentation.

\begin{table}[H]
    \begin{center}
        \begin{tabular}{|p{2.2cm}|R{1.3cm}|R{1.3cm}|R{1.4cm}|}
            \hline
        \textbf{} & \textbf{Num. comments} & \textbf{Avg. word length} & \textbf{Vocab. size} \\
        \hline
        CLEAN & 18,614 & 19.3 & 360,958 \\
        \hline
        OFFENSIVE & 13,823 & 11.3 & 157,517   \\
        \hline
        HATE & 11,051 & 23.6 & 260,841 \\
        \hline
        Total & 43,488 & 17.9 & 779,316 \\
        \hline
        \end{tabular}
    \end{center}
    \caption{The augmented HSD-VLSP courpus}
    \label{tab:dataset_augmented}
\end{table}

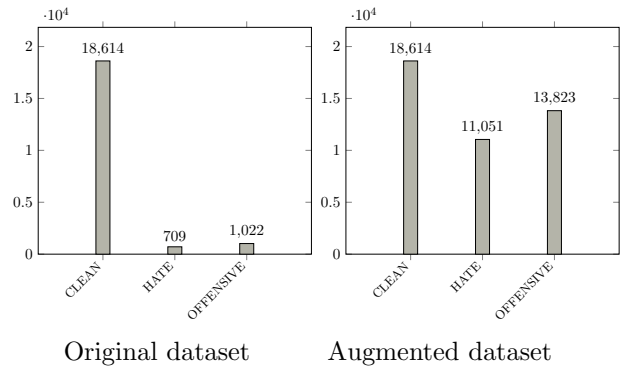
\begin{figure}[H]
\begin{minipage}[t]{.24\textwidth}
    \definecolor{grey}{HTML}{b4b4a9}
    \begin{tikzpicture}[scale=0.53]
        \begin{axis}[
            ybar,
            enlarge y limits={0.15,upper},
            enlarge x limits=0.45,
            symbolic x coords={CLEAN,HATE,OFFENSIVE},
            xtick=data,
            ymin = 0, ymax = 19000,
            nodes near coords,
            nodes near coords align={vertical},
    	    ylabel near ticks,
    	    x tick label style={font=\footnotesize,rotate=45, anchor=east}
        ]
        \addplot[black, fill=grey] coordinates {(CLEAN,18614) (HATE,709) (OFFENSIVE,1022)};
        \end{axis}
        
    \end{tikzpicture}
    \small \centering
    Original dataset
\end{minipage}
\begin{minipage}[t]{.2\textwidth}
    \definecolor{grey}{HTML}{b4b4a9}
    \begin{tikzpicture}[scale=0.53]
        \begin{axis}[
            ybar,
            enlarge y limits={0.15,upper},
            enlarge x limits=0.45,
            symbolic x coords={CLEAN,HATE,OFFENSIVE},
            xtick=data,
            ymin = 0, ymax = 19000,
            nodes near coords,
            nodes near coords align={vertical},
    	    ylabel near ticks,
    	    x tick label style={font=\footnotesize,rotate=45, anchor=east}
        ]
        \addplot[black, fill=grey] coordinates {(CLEAN,18614) (HATE,11051) (OFFENSIVE, 13823)};
        \end{axis}
    \end{tikzpicture}
    \small \centering
    Augmented dataset
\end{minipage}
    \caption{Number of comments on before and after augmentation in the HSD-VLSP dataset} 
    \label{fig:class_distribution_compare}
\end{figure}

It can be inferred from Table \ref{tab:dataset_augmented} that after applying the EDA techniques, the number of data and the vocabulary size on the HATE and OFFENSIVE labels and increased significantly (Words calculate the vocabulary size, and we use the pyvi\footnote{\url{https://pypi.org/project/pyvi/}} for tokenizing). Figure \ref{fig:class_distribution_compare} illustrates the distribution of three classes before and after using data augmentation techniques on the HSD-VLSP dataset. After using EDA techniques, the data on three labels are well-distributed. 

Besides, we apply the EDA on the UIT-VSFC training set to enhance the data on the neutral label for the sentiment-based task, and on \textit{facility} and \textit{other} labels for the topic-based task. Table \ref{tab:vsfc_overview_train_augmented} describes the UIT-VSFC training set after enhanced. Comparing with the original dataset as described in Table \ref{tab:vsfc_overview_train}, the number of comments and the vocabulary size of the neutral label on the sentiment-based task increased significantly. Same as the sentiment-based task, the number of comments and vocabulary size of the \textit{facility} and \textit{other} labels are also dramatically increased.

\begin{table}[H]
    \begin{center}
        \begin{tabular}{|p{2.1cm}|R{1.3cm}|R{1.3cm}|R{1.4cm}|}
        \hline
        \textbf{} & \textbf{Num. comments} & \textbf{Avg. word length} & \textbf{Vocab. size} \\
        \hline
        \multicolumn{4}{|c|}{\textbf{Sentiment-based task}}\\
        \hline
        Positive & 5,643 & 8.2 & 46,807 \\
        \hline
        Negative & 5,325 & 12.6 & 67,193  \\
        \hline
        Neutral & 4,697 & 8.1 & 38,349 \\
        \hline
        Total & 15,665 & 9.7 & 152,349 \\
        \hline
        \multicolumn{4}{|c|}{\textbf{Topic-based task}} \\
        \hline
        Lecturer & 8,166 & 9.7 & 79,854 \\
        \hline
        Training program & 2,201 & 12.2 & 27,039 \\
        \hline
        Facility & 5,906 & 13.7 & 81,299 \\
        \hline
        Others & 6,107 & 13.3 & 54,722 \\
        \hline
        Total & 22,380 & 10.8 & 242,914 \\
        \hline
        \end{tabular}
    \end{center}
    \caption{The augmented UIT-VSFC training set}
    \label{tab:vsfc_overview_train_augmented}
\end{table}

\begin{figure}[H]
    \begin{minipage}[t]{.24\textwidth}
        \definecolor{grey}{HTML}{b4b4a9}
        \begin{tikzpicture}[scale=0.54]
            \begin{axis}[
                ybar,
                enlarge y limits={0.15,upper},
                enlarge x limits=0.45,
                symbolic x coords={Positive,Negative,Neutral},
                xtick=data,
                ymin = 0, ymax = 6000,
                nodes near coords,
                nodes near coords align={vertical},
        	    ylabel near ticks,
        	    yticklabels=\empty,
        	    x tick label
        	    style={font=\footnotesize,rotate=45, anchor=east}
            ]
            \addplot[black, fill=grey] coordinates {(Positive,5643) (Negative,5325) (Neutral,458)};
            \end{axis}
        \end{tikzpicture}
        
        \small \centering
        Original training set
    \end{minipage}
    \begin{minipage}[t]{.23\textwidth}
        \definecolor{grey}{HTML}{b4b4a9}
        \begin{tikzpicture}[scale=0.54]
            \begin{axis}[
                ybar,
                enlarge y limits={0.15,upper},
                enlarge x limits=0.45,
                symbolic x coords={Positive,Negative,Neutral},
                xtick=data,
                ymin = 0, ymax = 6000,
                nodes near coords,
                nodes near coords align={vertical},
        	    ylabel near ticks,
        	    yticklabels=\empty,
        	    x tick label style={font=\footnotesize,rotate=45, anchor=east}
            ]
            \addplot[black, fill=grey] coordinates {(Positive,5643) (Negative,5325) (Neutral,4697)};
            \end{axis}
        \end{tikzpicture}
        \small \centering
        Augmented training set
    \end{minipage}
    \caption{The distribution of the sentiment-based task's labels of the UIT-VSFC dataset before and after enhanced} 
    \label{fig:vsfc_sentiment_task_compare}
\end{figure}
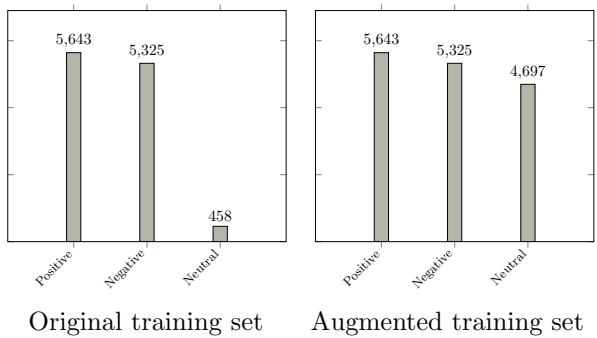

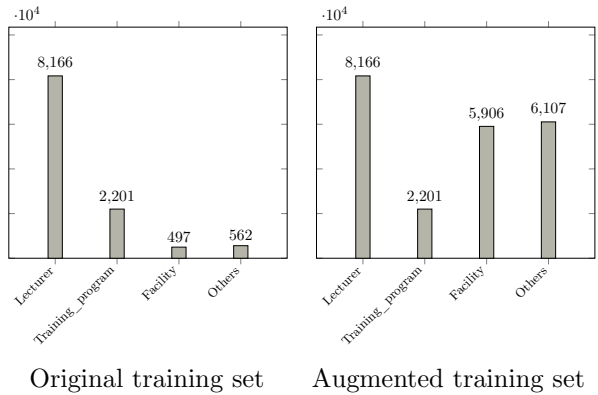
\begin{figure}[H]
    \begin{minipage}[t]{.24\textwidth}
        \definecolor{grey}{HTML}{b4b4a9}
        \begin{tikzpicture}[scale=0.54]
            \begin{axis}[
                ybar,
                enlarge y limits={0.15,upper},
                enlarge x limits=0.25,
                symbolic x coords={Lecturer,Training\_program,Facility,Others},
                xtick=data,
                ymin = 0, ymax = 9000,
                nodes near coords,
                nodes near coords align={vertical},
        	    ylabel near ticks,
        	    yticklabels=\empty,
        	    x tick label
        	    style={font=\footnotesize,rotate=45, anchor=east}
            ]
            \addplot[black, fill=grey] coordinates {
            (Lecturer,8166) 
            (Training\_program,2201) 
            (Facility,497) 
            (Others,562)
            };
            \end{axis}
        \end{tikzpicture}
        
        \small \centering
        Original training set
    \end{minipage}
    \begin{minipage}[t]{.23\textwidth}
        \definecolor{grey}{HTML}{b4b4a9}
        \begin{tikzpicture}[scale=0.54]
            \begin{axis}[
                ybar,
                enlarge y limits={0.15,upper},
                enlarge x limits=0.25,
                symbolic x coords={
                Lecturer,Training\_program,Facility,Others},
                xtick=data,
                ymin = 0, ymax = 9000,
                nodes near coords,
                nodes near coords align={vertical},
        	    ylabel near ticks,
        	    yticklabels=\empty,
        	    x tick label style={font=\footnotesize,rotate=45, anchor=east}
            ]
            \addplot[black, fill=grey] coordinates {
                (Lecturer,8166) 
                (Training\_program,2201) 
                (Facility,5906) 
                (Others,6107)
            };
            \end{axis}
        \end{tikzpicture}
        \small \centering
        Augmented training set
    \end{minipage}
    \caption{The distribution of the topic-based task's labels of the UIT-VSFC dataset before and after enhanced} 
    \label{fig:vsfc_topic_task_compare}
\end{figure}

Figure \ref{fig:vsfc_sentiment_task_compare} and Figure \ref{fig:vsfc_topic_task_compare} illustrate the UIT-VSFC training dataset before and after enhanced data on sentiment based task and topic based task respectively. For the two tasks, after augmentation the distribution of data between labels are balanced. 

\subsection{Model performance results}
\begin{figure}[H]
    \begin{minipage}[t]{.45\textwidth}
        \includegraphics[scale=0.5]{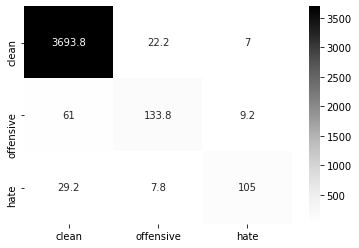}
        \\
        \centering
        The original dataset
    \end{minipage}
    
    \begin{minipage}[t]{.45\textwidth}
        \includegraphics[scale=0.5]{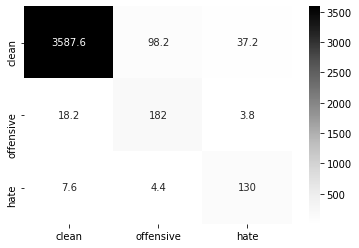}
        \\
        \centering
        The augmented dataset
    \end{minipage}
    \caption{Confusion matrix of Text-CNN model before and after enhanced data on the HSD-VLSP} 
    \label{fig:vlsp_confusion_matrix_compare}
\end{figure}

We implement the Text-CNN model on the entire original HSD-VLSP dataset and the augmented HSD-VLSP dataset. Table \ref{tab:empirical_resut} shows the result by F1-macro score. Comparing with the original results \cite{luu2020comparison}, the accuracy of the HSD-VSLP dataset after using augmented techniques are higher than the original dataset. According to Figure \ref{fig:vlsp_confusion_matrix_compare}, the number of right prediction on the \textit{offensive} and the \textit{hate} labels are increased.

\begin{table}[H]
    \begin{center}
        \begin{tabular}{|p{5cm}|R{2cm}|}
            \hline
        \textbf{Methodology} & \textbf{F1-macro (\%)}  \\
        \hline
        Text-CNN (original) \cite{luu2020comparison} &  83.04\\
        \hline
        \textbf{Text-CNN (augmented)} &  \textbf{84.80} \\
        \hline
        Maximum Entropy (original) \cite{luu2020comparison} &  64.58\\
        \hline
        \textbf{Maximum Entropy (augmented)} &  \textbf{75.27} \\
        \hline
        \end{tabular}
    \end{center}
    \caption{Empirical result by the Text-CNN model on the HSD-VLSP dataset}
    \label{tab:empirical_resut}
\end{table}

\begin{table}[H]
    \begin{center}
        \begin{tabular}{|L{4.1cm}|R{1.3cm}|R{1.3cm}|}
        \hline
        \textbf{Methodology} & \textbf{F1-micro (\%)}  & \textbf{F1-macro (\%)} \\
        \hline
        \multicolumn{3}{|c|}{\textbf{Sentiment-based task}}\\
        \hline
        Maximum Entropy (original) & 87.94 & 68.47 \\
        \hline
        Maximum Entropy (augmented) & \textbf{89.07} & \textbf{74.32} \\
        \hline
        Text-CNN (original) & \textbf{89.82} & 75.57 \\
        \hline
        Text-CNN (augmented) & 89.38 & \textbf{77.16} \\
        \hline
        \multicolumn{3}{|c|}{\textbf{Topic-based task}}\\
        \hline
        Maximum Entropy (original) & 84.03 & 71.23 \\
        \hline
        Maximum Entropy (augmented) & \textbf{86.03} & \textbf{74.87} \\
        \hline
        Text-CNN (original) & 86.63 & 75.23 \\
        \hline
        Text-CNN (augmented) & 86.32 & 74.86 \\
        \hline
        \end{tabular}
    \end{center}
    \caption{Empirical result of the UIT-VSFC dataset}
    \label{tab:vsfc_resut_train_augment}
\end{table}

Besides, Table \ref{tab:vsfc_resut_train_augment} shows the result of the UIT-VSFC dataset on the sentiment-based and the topic based tasks, respectively, before and after enhanced data. The original F1-micro score of the UIT-VSFC on both sentiment-based and topic-based tasks are referenced from \cite{8573337}. 

\begin{figure}[H]
    \begin{minipage}[t]{.45\textwidth}
        \includegraphics[scale=0.5]{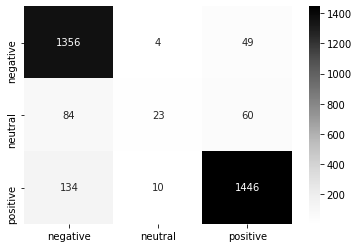}
        \\
        \centering
        The original training set
    \end{minipage}
    
    \begin{minipage}[t]{.45\textwidth}
        \includegraphics[scale=0.5]{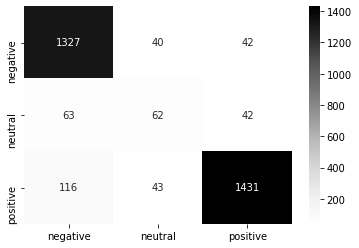}
        \centering
        \\
        The augmented training set
    \end{minipage}
    \caption{Confusion matrix of the Maximum Entropy model on the UIT-VSFC dataset before and after enhanced data for the sentiment-based task } 
    \label{fig:vsfc_confusion_matrix_compare_sent}
\end{figure}

According to Table \ref{tab:vsfc_resut_train_augment}, for the sentiment-based task, the UIT-VSFC dataset, after enhanced on the training set, gave better results than the original training set by the Maximum Entropy model on both F1-micro and F1-macro scores. The Text-CNN model gave better results by the F1-macro score when the training data are enhanced. For the topic based task, the result of Maximum Entropy are better after enhanced data. The Text-CNN results after augmented data, in contrast, are not as better as the original data. 

\begin{figure}[H]
    \begin{minipage}[t]{.45\textwidth}
        \definecolor{grey}{HTML}{b4b4a9}
        \includegraphics[scale=0.53]{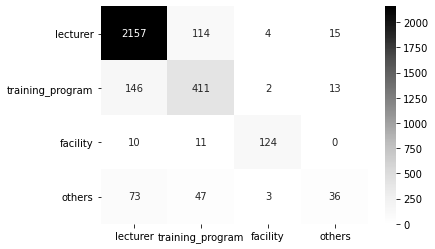}
        \centering
        The original training set
    \end{minipage}
    
    \begin{minipage}[t]{.45\textwidth}
        \includegraphics[scale=0.53]{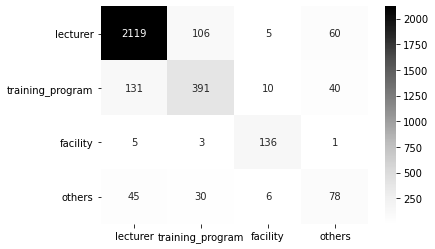}
        \centering
        The augmented training set
    \end{minipage}
    \caption{Confusion matrix of the Maximum Entropy model before and after enhanced data on the UIT-VSFC dataset for the topic-based task} 
    \label{fig:vsfc_confusion_matrix_compare_topic}
\end{figure}

In addition, Figure \ref{fig:vsfc_confusion_matrix_compare_sent} illustrates the confusion matrix of the UIT-VSFC dataset trained by the Maximum Entropy model before and after enhanced data for the sentiment-based task, and Figure \ref{fig:vsfc_confusion_matrix_compare_topic} indicates the confusion matrix of the UIT-VSFC dataset for the topic based task trained by the Maximum Entropy model. According to Figure \ref{fig:vsfc_confusion_matrix_compare_sent}, the ability of true prediction on the neutral label is increased after enhanced data. Nevertheless, according to Figure \ref{fig:vsfc_confusion_matrix_compare_topic}, the results before and after augmented data are just slightly different. Indeed, the enhanced data does not affect much on the performance result of the topic based task. 

Overall, for the HSD-VLSP hate speech dataset, the data augmentation techniques increase the models' performance. For the UIT-VSFC corpus, the data augmentation increased models' performance on the sentiment-based task by both Maximum Entropy and Text-CNN, while it does not impact the topic-based task.

\subsection{Error analysis}

According to Figure \ref{fig:vsfc_confusion_matrix_compare_topic}, on the UIT-VSFC dataset on the topic based task, the prediction of the  \textit{training\_program} label seems to be inclined to the \textit{lecture} label, and \textit{others} label seem to be inclined to the \textit{training\_program} and the \textit{lecturer} labels. Table \ref{tab:error_analysis_vsfc} listed examples of those cases. It can be inferred from Table \ref{tab:error_analysis_vsfc} that, most of cases the model predicted wrong to the \textit{lecture} label because the texts have words related to lecture such as: teacher, teaching, lesson, and knowledge. So does the \textit{training\_program} label with the appearance of words related to training program topic such as: subjects, requirements, and outcomes.
\begin{table}[H]
    \begin{center}
        \begin{tabular}{|p{0.55cm}|p{3cm}|R{1.1cm}|R{1.5cm}|}
            \hline
            \textbf{No.} & \textbf{Texts} & \textbf{True} & \textbf{Predict} \\
            \hline
            1 & cô nhiệt tình, giảng bài hiệu quả (English: The teacher is so enthusiastic and teaches very well) & 1 & 0 \\
            \hline
            2 & tiến độ dạy hơi nhanh (English: The teaching process is fast) & 1 & 0 \\
            \hline
            3 & sinh viên khó tiếp thu kiến thức (English: Student feel difficult to understand the knowledge) & 3 & 0 \\
            \hline
            4 & các yêu cầu của môn cần ghi rõ (English: The subject's  requirements should be well described) & 3 & 1 \\
            \hline
        \end{tabular}
    \end{center}
    \caption{Error analysis in the test set of the UIT-VSFC dataset on topic-based task. Label description: 0 - lecturer, 1 - training program, 2 - facility, 3 - others}
    \label{tab:error_analysis_vsfc}
\end{table}

\section{Conclusion}
\label{conclusion}
The imbalance in the datasets impact the performance of the machine learning models. Therefore, this paper focuses on the techniques that decreased the skewed distribution in the dataset by enhancing minority classes' data. We implemented the EDA techniques on the VLSP hate speech and the UIT-VSFC datasets and studied data augmentation's effectiveness on the imbalanced dataset. The results show that, when the data on the minority labels are increased, the model's ability to predict those labels is higher. However, the data augmentation techniques pull down the accuracy of other labels. Therefore, it is necessary to consider whether it is appropriate to apply the data augmentation techniques in a specific problem. 

In the future, we will construct the lexicon-based dictionary for sentiment analysis in the Vietnamese language, especially the abusive lexicon-based words like Hurtlex \cite{Bassignana2018HurtlexAM} for hate speech detection to improve the ability of the machine learning model. We will also implement modern techniques in text classification such as the BERT model \cite{devlin-etal-2019-bert} and the attention model \cite{yang-etal-2016-hierarchical} to increase the performance. Furthermore, in the hate speech detection problem, we will construct a new dataset which is more diverse in data sources and more balanced among classes.

\section*{Acknowledgments}
We would like to give our great thanks to the 2019 VLSP Shared Task organizers for providing a very valuable corpus for our experiments.

\bibliographystyle{acl}
\bibliography{reference}

\end{document}